\documentclass[runningheads]{llncs} 

\usepackage{cite}
\usepackage{amsmath,amssymb,amsfonts}
\usepackage{algorithmic}
\usepackage[ruled]{algorithm2e}
\usepackage{graphicx}
\usepackage{textcomp}
\usepackage{setspace} 
\usepackage{xcolor, soul}
\usepackage{multirow}

\usepackage{amsmath} 

\title{\LARGE \bf Towards Automated Differential Diagnosis of Skin Diseases Using Deep Learning and Imbalance-Aware Strategies}
\author{Ali Anaissi$^{1,2}$,  Ali Braytee$^{1}$, Weidong Huang$^{1}$, Junaid Akram$^{2}$, Alaa Farhat$^{1}$ and Jie Hua$^{3}$}

\institute{University of Technology Sydney, Australia \and
University of Sydney, Australia \and 
Shaoyang University, China  \\
\email{ali.anaissi@uts.edu.au,  ali.braytee@uts.edu.au, weidong.huang@uts.edu.au, Junaid.Akram@uts.edu.au,  aaf261@student.bau.edu.lb steven.hua@mq.edu.au}
}

\begin{document}
\maketitle
\thispagestyle{empty}
\pagestyle{empty}

\begin{abstract}
As dermatological conditions become increasingly common and the availability of dermatologists remains limited, there is a growing need for intelligent tools to support both patients and clinicians in the timely and accurate diagnosis of skin diseases. In this project, we developed a deep learning-based model for the classification and diagnosis of skin conditions. By leveraging pretraining on publicly available skin disease image datasets, our model was able to effectively extract visual features and accurately classify various dermatological cases.

Throughout the project, we refined the model architecture, optimized data preprocessing workflows, and applied targeted data augmentation techniques to improve overall performance. The final model, based on the Swin Transformer, achieved a prediction accuracy of \textbf{87.71\%} across eight skin lesion classes on the ISIC2019 dataset. These results demonstrate the model’s potential as a reliable diagnostic support tool for clinicians and a self-assessment aid for patients.

\end{abstract}

\section{Introduction}
\label{sec:introduction}

Skin diseases, particularly skin cancer, have become a significant global health concern. Early detection and accurate diagnosis are critical for effective treatment and positive patient outcomes. However, access to dermatologists remains limited worldwide, especially in regions where skin disease diagnosis is primarily handled by general practitioners \cite{Liu2020, anaissi2012dimension, anaissi2015case, zhou2022vgg}. Due to time constraints, limited resources, and a lack of specialized training, diagnostic accuracy by general practitioners can be suboptimal.

As a result, the use of advanced technologies, especially deep learning, has emerged as a promising solution to support both patients and clinicians in making faster and more accurate diagnostic decisions. In this context, the goal of our project is to apply deep learning techniques to improve skin disease classification, particularly under constraints such as limited image data and class imbalance. By incorporating domain knowledge with state-of-the-art deep learning methods, we aim to overcome current limitations and enhance diagnostic accuracy, ultimately contributing to better healthcare outcomes.

In this project, we used the widely adopted ISIC-2019 dermatology dataset and explored innovations in both model architecture and data augmentation. For deep learning models, we implemented the Swin Transformer, traditional CNNs, and ensemble CNN models. On the data augmentation side, we applied techniques such as SAM, AutoAugment, elastic deformation, and Fourier transforms to improve the model's ability to extract meaningful features from images.

Our best performance was achieved using the Swin Transformer in combination with these data enhancement techniques and optimized loss functions, resulting in an accuracy of 87.71\% on the classification task. This result demonstrates significant potential for supporting physicians in clinical settings by improving diagnostic precision and assisting in early detection of skin diseases.

\section{Related work}
\label{sec:related_work}
Dermatological classification has become a key area of research due to the increasing need for accurate and efficient skin disease diagnosis \cite{wu2025simplified, zhou2022vgg}. In recent years, several large datasets have been established to support the development of deep learning models for this purpose, with the ISIC series, SD-198, and HAM10000 being the most widely used. Among these, the ISIC-2019 dataset has gained particular prominence, validated by numerous studies for its robust and comprehensive data. For this research, the ISIC-2019 dataset has been selected as the target dataset, given its broad adoption and reliability in the field.

Over the past few years, a variety of deep learning models have been applied to dermatological image classification, with convolutional neural networks (CNNs), Transformer-based models, and multi-model integration approaches leading the way. CNNs, such as ResNet, DenseNet, and MobileNetV2, have consistently shown strong performance in image classification tasks. For instance, Rodrigues et al. \cite{Rodrigues2020} introduced a skin lesion classification method combining transfer learning and Internet of Things (IoT) systems, leveraging CNNs to improve diagnostic accuracy. Similarly, Gessert et al. \cite{Gessert2020} applied transfer learning to pre-trained CNN models on the ISIC series, achieving an impressive AUC of 97.9\%. ResNet models have also been tested extensively, with ResNet152 demonstrating the best performance in studies by Tan et al. \cite{Tan2023} and Mishra et al. \cite{Mishra2019}.

In more recent advancements, Transformer models have gained significant attention in dermatological classification. Liu et al. \cite{Liu2021} introduced the Swin Transformer, which employs a hierarchical structure with shifted windows to improve computational efficiency and capture a broader context. This method has shown substantial promise in enhancing the performance of dermatological image classification tasks. Zhou et al. \cite{Zhou2023} further advanced this by integrating the Vision Transformer (ViT) backbone with specialized modules to improve diagnostic accuracy, achieving a notable 74.5\% accuracy in clinical skin image diagnoses. Additionally, multimodal approaches that combine macro and micro images, as explored by Zhang et al.\cite{Zhang2024}, have also proven to enhance diagnostic efficiency.

In addition to well-established models, other mainstream approaches have contributed to dermatological classification. Srinivasu et al. \cite{AlTuwaijari2023} integrated MobileNetV2 with a long short-term memory (LSTM) structure to improve model design. This combined approach achieved an impressive accuracy of 87.17\%, showcasing excellent performance in skin lesion classification. Similarly, Groh et al. \cite{Groh2024} developed a pre-trained CVGG model tailored to assist physicians by automating parts of the classification system, thereby enhancing the decision-making process and solidifying the role of deep learning in clinical dermatological diagnoses.

Beyond individual model approaches, multi-model integration has become increasingly popular. For example, Alshahrani et al. \cite{Alshahrani2024} combined the features of DenseNet121, MobileNet, and VGG19 into a high-dimensional eigenmatrix, achieving an accuracy of 88.79\% after applying dimension reduction through t-SNE. Similarly, Gessert et al. \cite{Gessert2020} used a combination of EfficientNet, SENet, and ResNeXt models with soft voting to further improve classification accuracy and robustness.

One of the ongoing challenges in dermatological classification is the issue of class imbalance in datasets, which can hinder model performance. Yao et al. \cite{Yao2021} addressed this by developing the MWNL (Modified Weighted Negative Loss) method, which assigns higher loss weights to underrepresented categories, thereby improving the model's attention to difficult or rare classes. This approach significantly reduces the impact of class imbalance, providing a more balanced and accurate model for skin disease diagnosis.

Overall, the integration of state-of-the-art deep learning models, large datasets, and innovative approaches like multi-model integration and class imbalance mitigation has propelled the field of dermatological classification forward. Continued advancements in these areas will be critical for improving diagnostic accuracy and accessibility in skin disease detection.

\section{Proposed Framework}
\label{methods}
To address the challenges of imbalanced skin lesion datasets and enhance classification performance, we propose a hybrid framework built upon the Swin Transformer architecture pretrained on ImageNet-1K. While this pretrained backbone provides a strong starting point, domain-specific adaptation is necessary to improve generalizability across underrepresented disease categories.

\begin{figure}[!t] 
\begin{center} 
\includegraphics[scale=0.3]{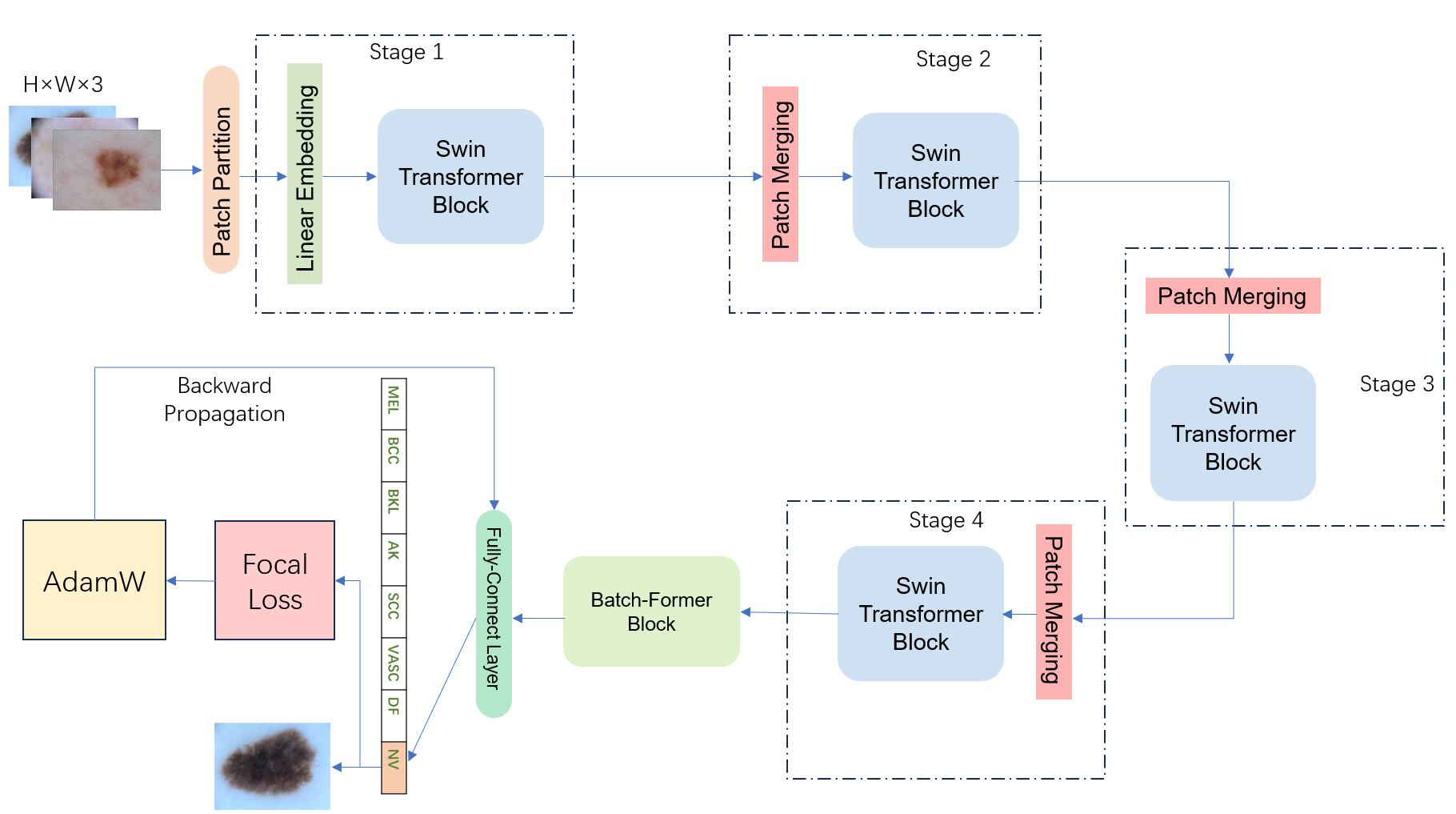} 
\end{center} 
\caption{Overview of the proposed model structure based on Swin-Transformer.} \label{fig:model_structure} 
\end{figure}

To that end, we implement a series of architectural modifications and integrate multiple techniques—including BatchFormer, Focal Loss, and a dynamic learning rate scheduler (ReduceLROnPlateau)—to boost performance, particularly for tail classes. The model follows a four-stage hierarchical design, where each stage progressively reduces spatial resolution while increasing feature dimensionality, enabling extraction of increasingly abstract semantic features. The overall structure is illustrated in Figure \ref{fig:model_structure}.

\subsection{Swin Transformer}
The Swin Transformer, proposed by Ze Liu et al. \cite{Liu2021}, is a hierarchical vision transformer that improves computational efficiency through localized self-attention mechanisms. Unlike traditional Vision Transformers (ViTs) that operate over global attention and are computationally expensive, the Swin Transformer introduces two novel mechanisms—Shifted Windows (SW-MSA) and Patch Merging—to process images in a hierarchical and efficient manner.

Built upon the success of Transformers in NLP, Swin Transformer combines the benefits of self-attention with the hierarchical representation capabilities of CNNs. This makes it particularly suitable for tasks involving low-resolution and variable-sized medical images. Each Swin Transformer Block consists of two core components: Window-based Multi-head Self-Attention (W-MSA) and Shifted Window-based Multi-head Self-Attention (SW-MSA).

\begin{figure}[!t] 
\centering 
\includegraphics[scale=0.25]{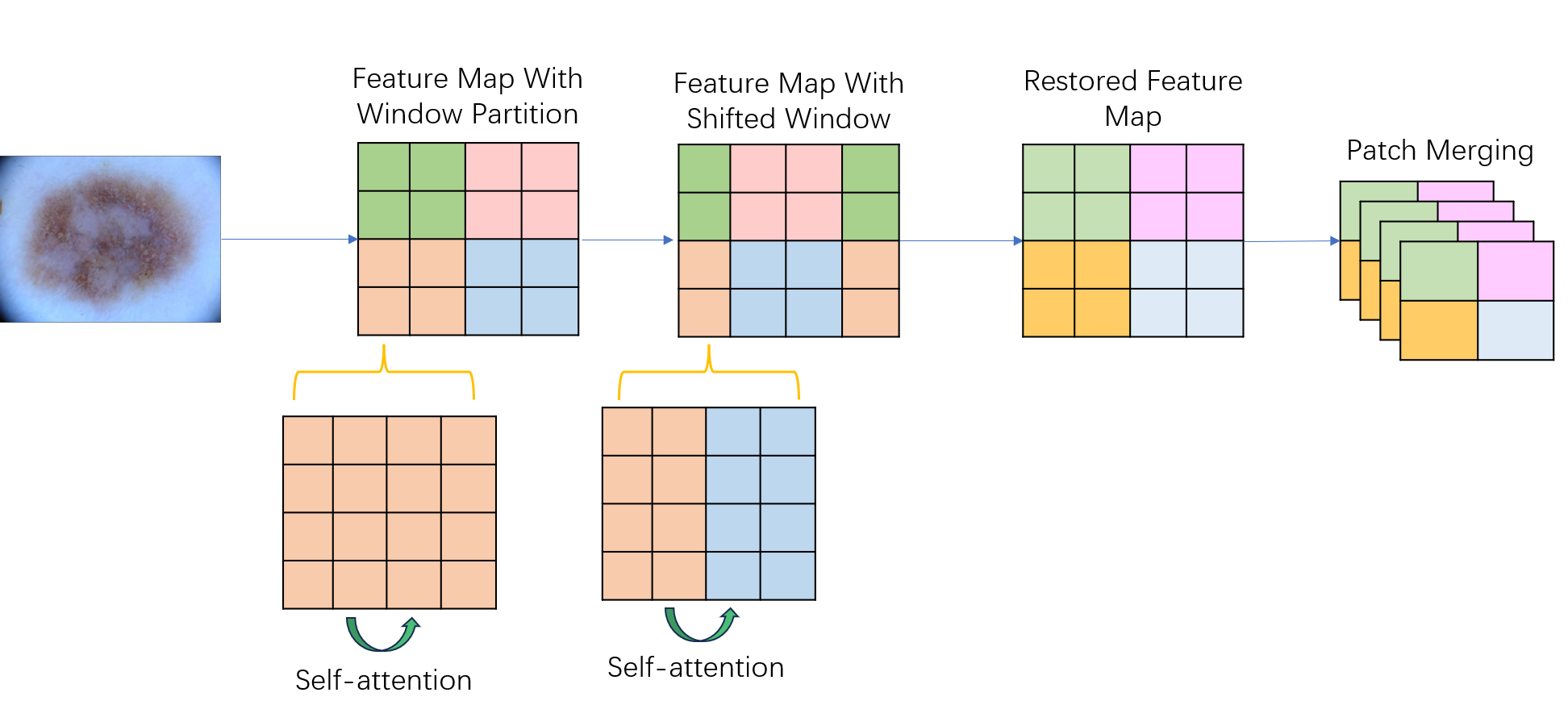} 
\caption{W-MSA to SW-MSA Attention Process in the Swin-Transformer Block.} 
\label{fig:st_process}
\end{figure}

W-MSA restricts attention computation within non-overlapping windows, while SW-MSA allows information exchange across windows by shifting the partitioned windows. This combination enables both local and global feature learning, improving the model’s ability to capture complex spatial relationships. The full process of transitioning from W-MSA to SW-MSA is illustrated in Figure \ref{fig:st_process}.

Patch merging is employed at each stage to reduce spatial resolution while increasing feature depth. For an input of size $(H, W, D)$, patch merging reduces the spatial dimensions to $(H/2, W/2)$ and quadruples the feature depth to $4D$, enabling the model to extract more complex features at lower resolutions while minimizing computational cost.

This hierarchical structure, combined with localized attention and channel dimension expansion, equips the model with strong representational capacity suitable for medical imaging tasks, including skin disease classification.

\subsection{BatchFormer}
To address the issue of class imbalance—particularly the under-representation of rare diseases—we incorporate BatchFormer \cite{Hou2022}, a module designed to enhance representation learning for tail classes by leveraging inter-sample relationships within a mini-batch.

\begin{figure}[!t] 
\centering 
\includegraphics[scale=0.3]{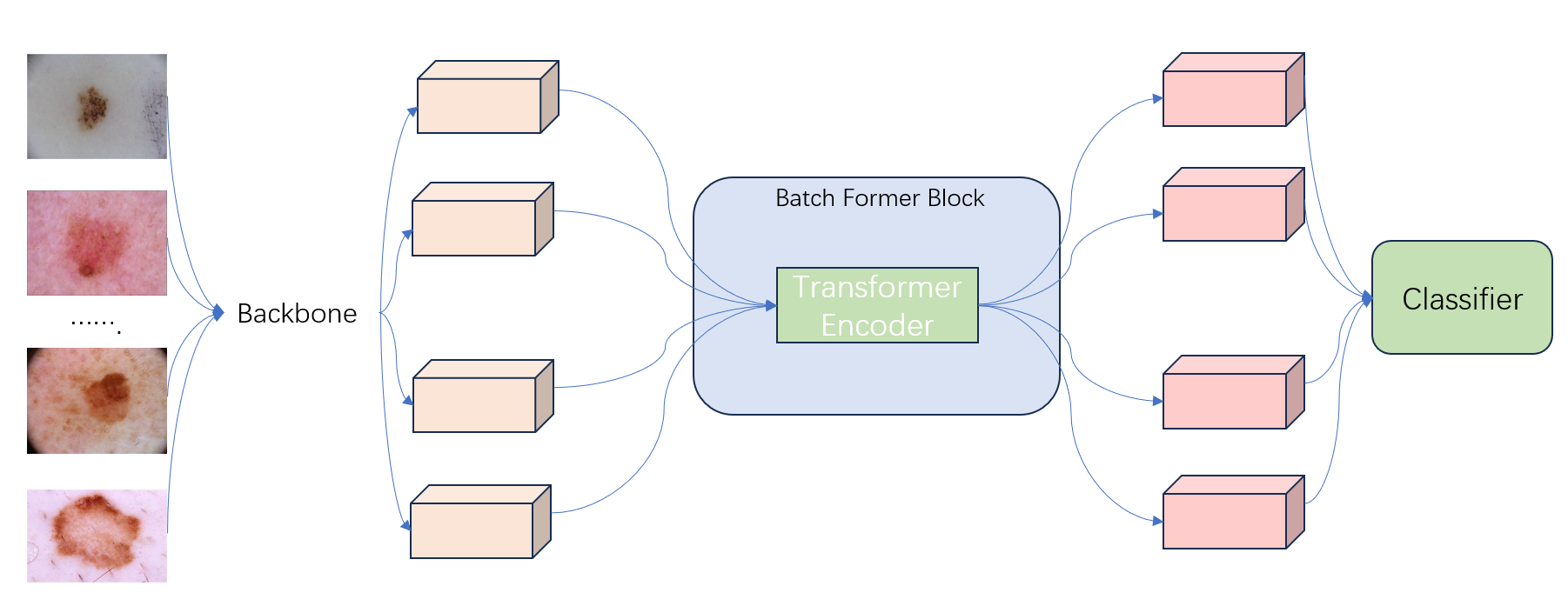} 
\caption{BatchFormer Block Architecture.}
\label{fig:bf_block} 
\end{figure}

BatchFormer applies a Transformer encoder across the batch dimension, encoding not only individual sample features but also the similarity between samples. The architecture of the BatchFormer block is illustrated in Figure \ref{fig:bf_block}.

\begin{figure}[!t] 
\centering 
\includegraphics[scale=1]{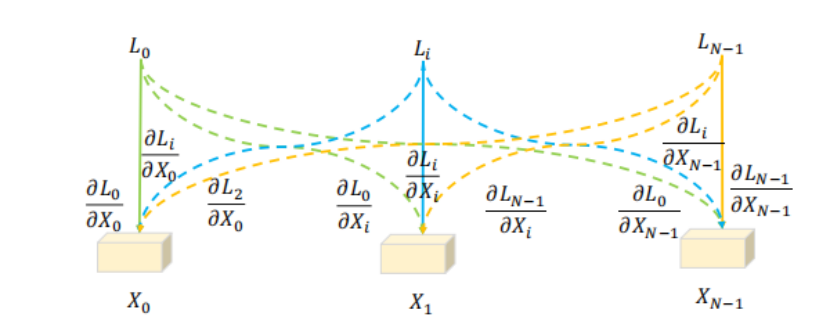} 
\caption{Cross-sample Gradient Propagation in BatchFormer.} 
\label{fig:cs_g_p}
\end{figure}

This structure enables cross-sample gradient propagation, wherein gradients for each sample are influenced by those of others in the batch. As illustrated in Figure \ref{fig:cs_g_p}, this mechanism allows knowledge transfer from head classes to tail classes, acting as a form of virtual data augmentation for underrepresented categories.

By aligning feature distributions between classes, BatchFormer helps mitigate bias toward majority classes and improves the model’s ability to generalize to minority classes \cite{Hou2022, anaissi2013balanced}.

This mechanism acts as a form of virtual data augmentation, enabling minority classes to benefit from rich gradients produced by majority samples, thereby improving classification performance across underrepresented classes.

\subsection{Focal Loss for Imbalanced Learning}

To further address the challenge of class imbalance in our dataset, we incorporate the Focal Loss function, originally proposed by Lin et al. \cite{Lin2017}. Unlike standard cross-entropy loss, Focal Loss dynamically scales the loss contribution from well-classified examples and focuses learning on hard misclassified instances. This is particularly beneficial in scenarios like medical image classification, where minority classes are often underrepresented and easily misclassified.

The Focal Loss introduces a modulating factor to the standard cross-entropy loss:

\begin{equation}
\mathcal{L}_{\text{focal}} = -\alpha_t (1 - p_t)^\gamma \log(p_t)
\end{equation}

where:
\begin{itemize}
    \item $p_t$ is the model’s estimated probability for the true class,
    \item $\alpha_t$ is a weighting factor for class $t$ that balances positive and negative examples,
    \item $\gamma$ is the focusing parameter that reduces the relative loss for well-classified examples.
\end{itemize}

When an example is correctly classified with high confidence (i.e., $p_t$ is large), the term $(1 - p_t)^\gamma$ becomes small, thereby reducing the loss contribution from that example. Conversely, for misclassified or low-confidence examples, the loss remains high, forcing the model to focus on those harder cases during training.

In our application, Focal Loss enables the model to prioritize learning from underrepresented and difficult examples, which helps improve the classification of minority classes while mitigating overfitting to the majority class. Unlike traditional sampling strategies or cost-sensitive learning, Focal Loss reshapes the loss landscape directly, making it computationally efficient and straightforward to integrate into modern deep learning frameworks.

\subsection{Learning Rate Scheduling with ReduceLROnPlateau}
To optimize training stability and convergence, we employ the ReduceLROnPlateau scheduler, which dynamically adjusts the learning rate based on the validation loss plateau. This scheduler reduces the learning rate when performance stagnates, allowing the model to fine-tune its parameters in later stages of training.

In combination with Focal Loss, this approach further enhances learning dynamics. Focal Loss encourages the model to focus on difficult samples, while ReduceLROnPlateau ensures that the learning rate is appropriately scaled to allow for fine-grained updates. Together, they help prevent overfitting, improve convergence speed, and boost classification performance on imbalanced datasets.

\section{Experiments}
\subsection{Dataset Description and Preparation}

In this study, we use the ISIC2019 dataset, a publicly available benchmark provided by the International Skin Imaging Collaboration (ISIC). The data set comprises thousands of dermoscopic images labeled in various categories of skin lesion and is accessible through the official ISIC archive.

\begin{figure}[!t]
\centering
\includegraphics[scale=0.25]{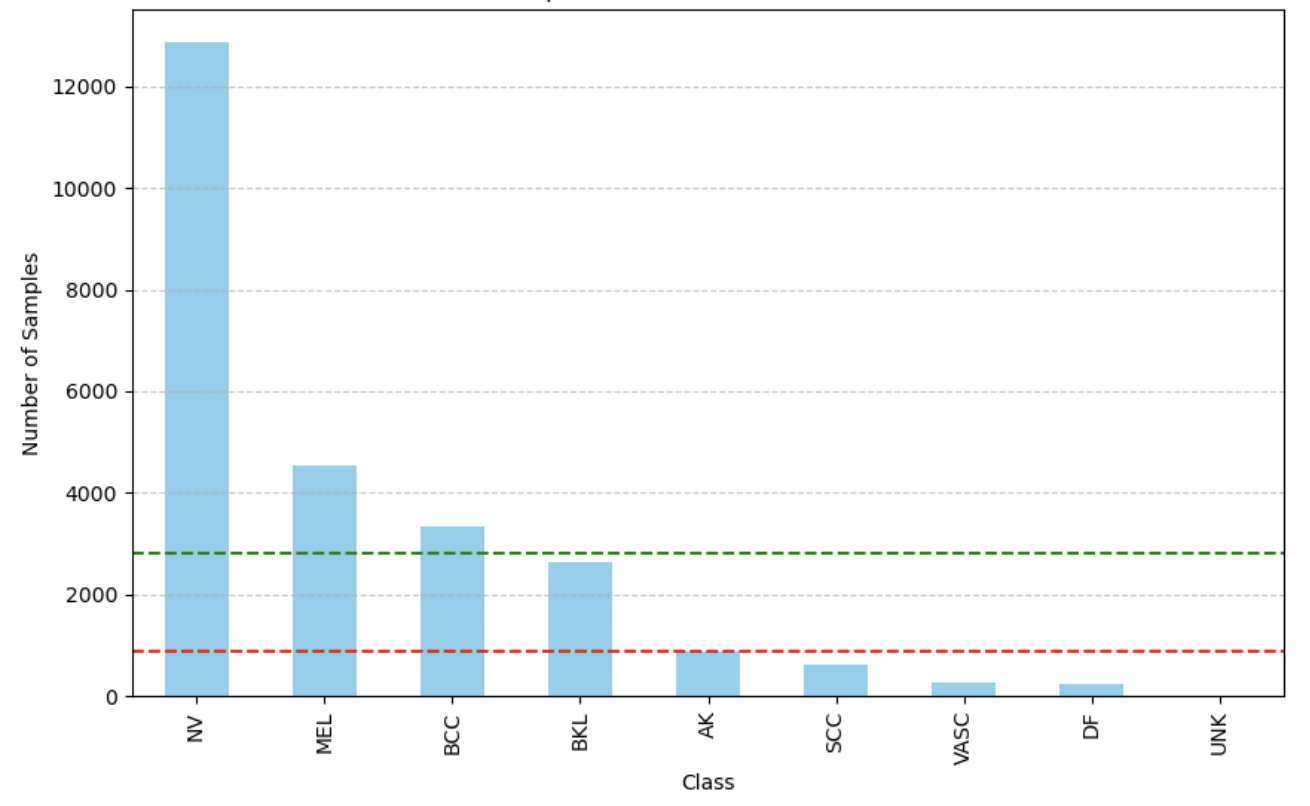}
\caption{Class distribution in the ISIC2019 dataset.}
\label{fig:isic2019}
\end{figure}

To gain insight into the dataset and identify potential classification challenges, we performed a preliminary statistical analysis of class distributions. Using \texttt{NumPy}, we computed the number of samples in each category, along with their corresponding mean and median. The analysis, visualized in Figure~\ref{fig:isic2019}, reveals a severe class imbalance. For example, the “NV” (melanocytic nevus) class comprises more than 12,000 samples, while the “DF” (dermatofibroma) class has fewer than 300. This uneven distribution presents a significant challenge in training robust and fair classifiers. Consequently, our method incorporates several imbalance-aware learning strategies, including Focal Loss and BatchFormer.

To prepare the dataset for training and evaluation, we split it into three subsets: 70\% for training, 15\% for validation, and 15\% for testing. We adopted a stratified sampling strategy to ensure that the class distribution remains consistent across all subsets, preserving the imbalance characteristics while enabling fair model evaluation.

\begin{figure}[!t]
\centering
\includegraphics[scale=0.25]{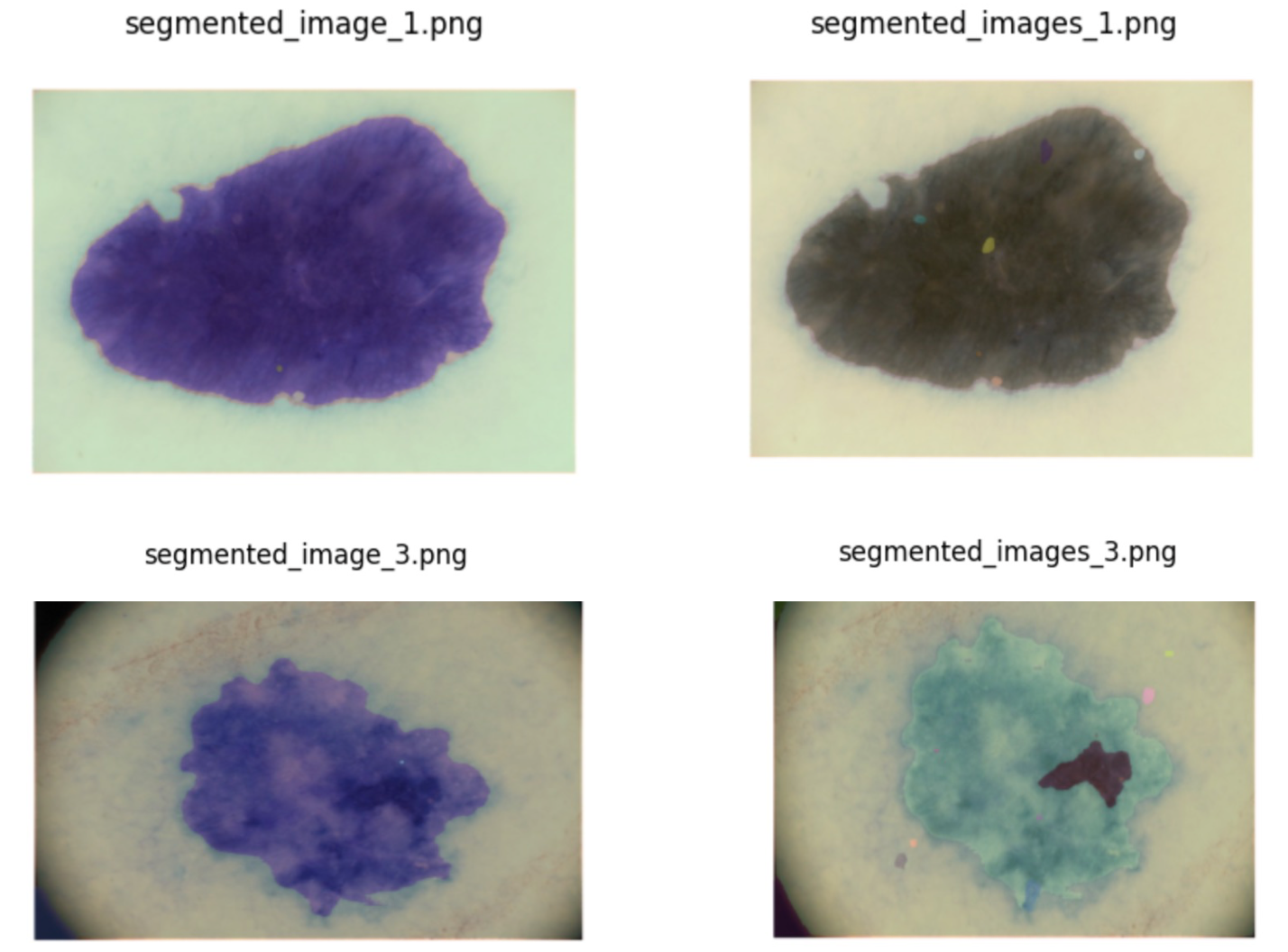}
\caption{Examples of lesion masks generated using SAM.}
\label{fig:sam_example}
\end{figure}

\begin{table}[!t] 
\centering 
\caption{Classification accuracy with and without SAM preprocessing.} 

\begin{tabular}{|c|c|} 
\hline Model & Accuracy \\
\hline ViT & 65.6\% \\ 
\hline ViT + SAM & 62.0\% \\
\hline ResNet & 75.6\% \\
\hline ResNet + SAM & 73.0\% \\
\hline Swin Transformer & 76.6\% \\ 
\hline Swin Transformer + SAM & 70.8\% \\
\hline 
\end{tabular}
\label{tab:acc_sam} 
\end{table}

\begin{table}[!t]
\centering 
\caption{Classification accuracy with data augmentation.}
\small 
\begin{tabular}{|c|c|} 
\hline Model & Accuracy \\
\hline DenseNet & 71.53\% \\
\hline DenseNet + AutoAugment & 73.87\% \\
\hline Swin Transformer & 87.34\% \\
\hline Swin Transformer + Elastic Deformation & 87.71\% \\
\hline 
\end{tabular} 
\label{tab:densenet&st&aug} 
\end{table}

\begin{table*}[!t] 
\small 
\centering 
\caption{Model performance comparison on ISIC2019.} 
\begin{tabular}{|c|c|c|} 
\hline Model Category & Model & Accuracy \\
\hline \multirow{9}{*}{CNN} & DenseNet121-CE & 76.47\% \\
\cline{2-3} & DenseNet121-MWNL & 61.37\% \\
\cline{2-3} & DenseNet121 + Fourier Transformer & 80.16\% \\
\cline{2-3} & EfficientNet & 80.94\% \\
\cline{2-3} & Ensemble (Weighted) & 73.18\% \\
\cline{2-3} & Ensemble (Concatenated) & 68.89\% \\ 
\cline{2-3} & U-Net & 60.52\% \\
\cline{2-3} & ResNet50 + SimAM & 69.18\% \\
\hline \multirow{2}{*}{Transformer} & ViT & 80.16\% \\
\cline{2-3} & \textbf{Swin Transformer + BatchFormer + Focal Loss + ReduceLROnPlateau} & \textbf{87.71\%} \\
\hline
\end{tabular} 
\label{tab:acc_compare} 
\end{table*}

\subsection{Experimental Setup}

\subsubsection{Data Preprocessing and Augmentation}
To mitigate the effects of class imbalance, we implemented multiple data preprocessing and augmentation strategies. Initially, we explored the use of the Segment Anything Model (SAM), developed by Meta \cite{Kirillov2023}, to generate segmentation masks for each lesion. These masks were used to render the lesion regions more distinctly by emphasizing their boundaries. The hypothesis was that focusing attention on these segmented regions could improve classification accuracy by highlighting lesion-relevant features. In addition to segmentation, we applied two augmentation techniques:
\begin{itemize}
    \item Elastic Deformation, which distorts images using Gaussian-based displacement fields to simulate variability in lesion shapes \cite{Chlap2021}.

    \item AutoAugment, a policy-based augmentation method that applies combinations of transformations (e.g., rotation, color jitter) to enrich the training set.
\end{itemize}

Elastic Deformation was applied selectively to disease classes with fewer than 2,000 training samples in order to directly address the imbalance.

\section{Results and Discussion}

Our study aimed to enhance skin lesion classification by incorporating advanced data preprocessing, augmentation, and optimization strategies within a Transformer-based deep learning framework. The proposed system comprises two core components: (1) a preprocessing and augmentation pipeline tailored to address dataset-specific challenges, and (2) a classification model based on the Swin Transformer, augmented with optimization modules to improve performance on imbalanced data.

To evaluate the contributions of each component, we conducted experiments in three phases: assessing segmentation-based preprocessing, applying data augmentation techniques, and performing model-level optimization. The findings from each phase are discussed below.

\subsection{Impact of Segmentation-Based Preprocessing}
We initially investigated the impact of segmentation as a preprocessing step, using the Segment Anything Model (SAM) to generate lesion masks. These masks were used to isolate lesion boundaries in dermoscopic images with the hypothesis that enhancing these regions would assist the classifier in focusing on diagnostically relevant areas.

However, our empirical findings contradicted this expectation. Table~\ref{tab:acc_sam} shows a consistent drop in accuracy across all tested architectures (ViT, ResNet, and Swin Transformer) when SAM preprocessing was applied. This suggests that the segmentation process might introduce artifacts or omit subtle yet critical lesion characteristics such as color variegation, asymmetric texture, and pigment networks—features that are vital in dermatological assessments.

Moreover, SAM, while powerful, is a general-purpose model not specifically fine-tuned for dermoscopic images. This limitation may lead to inaccurate segmentation boundaries, especially for complex or irregularly shaped lesions, thus introducing noise rather than improving the focus of the classifier. These observations highlight the importance of using domain-specific segmentation tools or integrating lesion-aware attention mechanisms instead of general segmentation as a preprocessing step.

\subsection{Effectiveness of Data Augmentation}
To address class imbalance and enrich feature variability, we incorporated two data augmentation techniques: AutoAugment and Elastic Deformation. AutoAugment applies diverse transformation policies (e.g., rotation, brightness adjustment, shear), thereby exposing the model to a broader range of intra-class variations. Elastic Deformation introduces spatial distortions that simulate natural biological variation in lesion morphology.

Table~\ref{tab:densenet&st&aug} shows that both augmentation techniques yielded performance improvements across different models. For instance, DenseNet benefited from a 2.3\% accuracy gain with AutoAugment, while Swin Transformer achieved the highest accuracy of 87.71\% when augmented with Elastic Deformation. Notably, we applied Elastic Deformation selectively to underrepresented classes, which suggests that its targeted use contributes effectively to rebalancing class representation during training.

These results affirm that thoughtful augmentation not only mitigates class imbalance but also strengthens model robustness by enabling better generalization across diverse lesion appearances. This is especially crucial in medical imaging, where real-world variability is high and overfitting to limited visual cues can impair diagnostic utility.

\subsection{Model Comparison and Final Results}
Building upon the improvements introduced through augmentation, we proceeded to compare model architectures and integrate optimization strategies to further refine classification performance. A diverse range of models, including traditional CNNs and modern Transformer-based architectures, were evaluated on the ISIC2019 dataset. Table~\ref{tab:acc_compare} summarizes the results.

CNN-based models such as DenseNet121 and EfficientNet performed competitively, with EfficientNet reaching an accuracy of 80.94\%. Incorporating techniques like Fourier Transformer and attention modules further boosted CNN performance. However, Transformer-based models demonstrated superior accuracy, likely due to their enhanced capability to model long-range dependencies and global patterns.

The best-performing configuration combined Swin Transformer with BatchFormer, Focal Loss, and ReduceLROnPlateau, achieving a classification accuracy of 87.71\%. The Swin Transformer’s hierarchical design and shifted window attention mechanism enabled efficient multi-scale feature extraction—critical for analyzing both fine lesion textures and broader shape asymmetries. BatchFormer facilitated inter-sample learning by encouraging representation sharing among samples in the same mini-batch, effectively improving minority class recognition.

Focal Loss was instrumental in mitigating the impact of class imbalance by down-weighting easy samples and emphasizing harder, misclassified ones during training. Lastly, ReduceLROnPlateau ensured smoother convergence and prevented overfitting by dynamically adjusting the learning rate in response to validation performance.

Overall, these results validate the proposed approach as an effective solution for skin lesion classification, particularly in the context of class imbalance and heterogeneous data distributions. The integration of Transformer-based architectures with tailored augmentation and optimization strategies offers a promising direction for future work in medical image analysis.

\section{conclusion}
Our findings highlight the Swin Transformer as a highly capable architecture for skin lesion classification, particularly when working with large-scale datasets that exhibit significant class imbalance. Compared to conventional CNN-based models, the Swin Transformer consistently achieved superior results—most notably reaching an accuracy of 87.71\% when combined with BatchFormer, Focal Loss, and the ReduceLROnPlateau learning rate scheduler. This demonstrates its ability to effectively capture both local texture and global contextual information through its hierarchical self-attention mechanism.

However, despite this strong performance, the Swin Transformer does not include any built-in mechanism to directly handle data imbalance. Furthermore, the model was initialized with pretrained weights from general-purpose datasets such as ImageNet, which may not be optimally suited for the specific characteristics of dermoscopic images used in skin disease classification.

To further enhance performance, future research could focus on fine-tuning the model using domain-specific dermatology datasets to improve its sensitivity to lesion-specific features. Additionally, integrating more robust imbalance-aware learning strategies such as dynamic class-weighted loss functions or adaptive sampling techniques, may help improve classification performance for underrepresented lesion types and ensure more balanced predictions across all categories.

\section*{Acknowledgements}
We acknowledge each of these students — Xinyu Ma, Frank Cheng, Sean Xia, Huiwen Zhao, Yuhao Xiong and Yibo Ding for their hard work, resilience, and commitment throughout this research endeavor.

 \bibliographystyle{splncs04}
 \bibliography{ref}
 


\end{document}